\documentclass[11pt]{article}

\usepackage{fullpage}
\usepackage{graphicx}
\usepackage{subfigure}
\usepackage{algorithm}
\usepackage{algorithmic}
\usepackage{amsmath}
\usepackage{amsfonts}
\usepackage{amssymb}
\usepackage{url}
\usepackage{color}
\definecolor{greytext}{gray}{0.5}
\usepackage{lscape}
\usepackage{booktabs}
\usepackage{epstopdf}
\usepackage{subfig}
\usepackage{multirow}

\include{mydef}

\title{Decentralized, Adaptive, Look-Ahead Particle Filtering}
\author{Mohamed Osama Ahmed\footnote{Authorship in alphabetical order.}, Pouyan T. Bibalan, Nando de Freitas and Simon Fauvel  \\
CS and ECE Departments \\
  University of British Columbia \\
  Vancouver, Canada \\
  \texttt{ \{mohameda,pouyant,simonf\}@ece.ubc.ca} \;\; and \texttt{ nando@cs.ubc.ca}
}
\date{\today}

\begin{document}

\maketitle

\begin{abstract}
The decentralized particle filter (DPF) was proposed recently to increase the level of parallelism of particle filtering. Given a decomposition of the state space into two nested sets of variables, the DPF uses a particle filter to sample the first set and then conditions on this sample to generate a set of samples for the second set of variables. The DPF can be understood as a variant of the popular Rao-Blackwellized particle filter (RBPF), where the second step is carried out using Monte Carlo approximations instead of analytical inference. As a result, the range of applications of the DPF is broader than the one for the RBPF. In this paper, we improve the DPF in two ways. First, we derive a Monte Carlo approximation of the optimal proposal distribution and, consequently, design and implement a more efficient look-ahead DPF. Although the decentralized filters were initially designed to capitalize on parallel implementation, we show that the look-ahead DPF can outperform the standard particle filter even on a single machine. Second, we propose the use of bandit algorithms to automatically configure the state space decomposition of the DPF. 
\end{abstract}

\section{Introduction}

Without a doubt, Rao-Blackwellization has proved to be the most successful technique for enabling particle filters to solve high-dimensional dynamic inference problems, see for example \cite{doucet2001particle,Doucet2000dbns,df2002fault,Montemerlo2002} and the many citations to those papers. When applying Rao-Blackwellization to particle filtering, one decomposes the state space into two groups of variables. The first group of variables is sampled with a particle filter. Then one conditions on these samples to compute the sufficient statistics of the second group of variables analytically. If a decomposition exists such that the dimension of the sampled variables is small while the dimension of the analytical variables is large, then one can effectively solve high dimensional problems. An example of this, of great practical relevance, is the application of RBPFs to jump-Markov linear Gaussian systems \cite{df2004fault,Schon2005,Caron2007,Liao2007learning}. 

These works have however left open some important questions: (i) What happens if there is no analytical expression for the distribution of the second group of variables? (ii) Is there a reason for deriving an approximate Rao-Blackwellized Particle Filter (RBPF) in this case? (iii) Instead of only two groups of variables, can one use successive nesting of more than two groups? (iv) How do we decompose the state space automatically?

Chen \emph{et. al.} (2011) \nocite{chen2011decentralized} have recently provided answers to questions (i) and (ii). They designed a new particle filter, which they named the Decentralized Particle Filter (DPF), that is effectively an RBPF, but with the difference that the distribution of the second group of variables is also approximated by a conditional particle filter. That is, one uses a particle filter to sample the first group and then conditions on these samples to sample the second group with a conditional particle filter. They provide an important reason for doing this: increased parallelization. 

The resampling step is one of the computational bottlenecks in parallel implementations of particle filters in graphics processing units (GPUs) and field-programmable gate arrays (FPGAs)  \cite{bolic2005resampling, m�guez2007analysis,Lee2010}. By decomposing the state space, the DPF allows for more efficient, local in the state space, resampling. Chen and colleagues have demonstrated this advantage of DPFs over standard PFs. Moreover, with increased interest in the deployment of particle filters for large scale applications, such as the analysis of streaming news \cite{Ahmed2011}, algorithms that capitalize on decompositions of the space space are of great research interest.   

Chen  \emph{et. al.} suggest the use of Gaussian approximations in order to manage computation. In this paper, we show that it is possible to avoid these Gaussian approximations without significant loss of performance. Moreover, we show that it is possible to obtain a pure Monte Carlo approximation of the optimal importance distribution (optimal proposal). This Monte Carlo approximation enables us to design and implement a
 look-ahead filter, where the sampling and resampling steps can be swapped. In the context of exact Rao-Blackwellization, this look-ahead strategy is described in detail in \cite{df2004fault} and was first suggested in \cite{doucet2001particle}. In \cite{df2004fault}, it was clear that the look-ahead RBPF performed significantly better than the PF and RBPF algorithms in practical domains. In our context, the derivation of a look-ahead strategy is a bit more tricky as it involves additional Monte Carlo approximations. However, as we will see in the experiments, the look-ahead strategy still results in substantial improvements over the standard DPF.

Our final contribution is to answer, to some extent and for the first time, question (iv). This question was posed more than ten years ago and continues appearing in the future work sections of papers on the topic, including the DPF paper. Our solution involves the usage of online bandit algorithms \cite{auer1995gambling} to decide the order in which variables should be sampled. Question (iii) is still open, but we conjecture that the improvements introduced in \cite{chen2011decentralized} and here will lead to it being answered in the near future.

The paper is organized as follows. Section 2 describes the models, poses the inference problems, and provides a brief description of the DPF. The section ends with a description of the proposed
 look ahead (LA)-DPF algorithm. Section 3 presents the automatic state decomposition strategy. 
The PF, DPF and LA-DPF are compared in the experiments of Section 4. We conclude the paper in Section 5.

\section{Decentralized Particle Filter}

The state space is decomposed into two groups of variables $(x_{t}  \in {\cal X} \subseteq \mathbb{R}^{n_{x}}, z_{t} \in {\cal Z} \subseteq \mathbb{R}^{n_{z}})$, which are governed by the following latent, dynamic state space model with observations $y_t  \in {\cal Y} \subseteq \mathbb{R}^{n_{y}}$:
\begin{eqnarray}\label{eq:mainDPF}
x_{t+1} &=& f_{t}^{x} (x_{t}, z_{t}, v_{t}^{x}) \nonumber\\
z_{t+1} &=& f_{t}^{z} (x_{t:t+1}, z_{t}, v_{t}^{z})  \\
y_{t} &=& h_{t}(x_{t}, z_{t}, e_{t}), \nonumber
\end{eqnarray}
where $x_{t:t+1}=(x_t,x_{t+1})$, $v_{t} = (v_{t}^{x} , v_{t}^{z})$ and $e_t$ are noise processes, and $f(\cdot)$ and $h(\cdot)$ are nonlinear mappings. This model can be equivalently expressed in terms of the initial distributions 
$p(x_0)$ and $p(z_0|x_0)$, the transition distributions $p(x_{t+1}|x_t,z_t)$ and $p(z_{t+1}|x_{t:t+1},z_t)$ and the observation model $p(y_t|x_t,z_t)$. We assume that the parameters of these distributions are known and focus on the inference problem.
\begin{figure}[t!]
\begin{center}
\includegraphics[scale=.9]{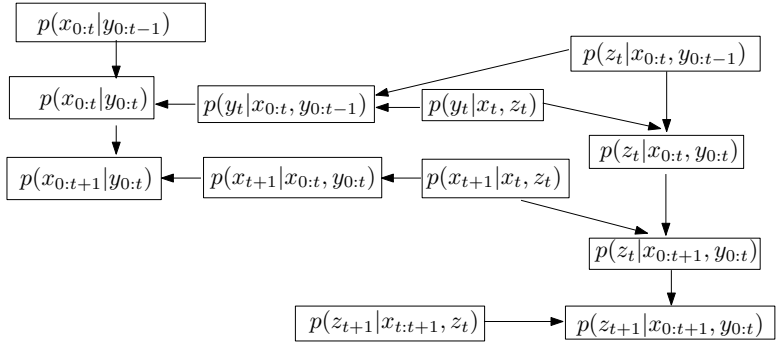}
\caption{{\small \it Flow of distributions that need to be computed in order to solve the nested filtering problem.}}
\label{fig:DPFsketch}
\end{center}
\end{figure}

The goal of inference is to recursively estimate the posterior distribution $p(z_t,x_{0:t}|y_{0:t})$. Using the following factorization:
\begin{equation}\label{eq:filteringProblem}
p(z_{t}, x_{0:t}| y_{0:t}) = p(z_{t}|x_{0:t}, y_{0:t}) p(x_{0:t}|y_{0:t}),
\end{equation}
with $x_{0:t} = (x_0,\dots,x_t)$, this filtering problem can be split into the two nested subproblems of recursively estimating (1) $p(x_{0:t}|y_{0:t})$ and (2) $p(z_{t}|x_{0:t}, y_{0:t})$. The way in which these subproblems interact with each other is depicted in Figure \ref{fig:DPFsketch}.
The diagram shows the necessary steps for implementing the optimal filter in this nested setting. However,
except in very specific cases, there is no analytic solution to this filtering problem. Therefore, a numerical algorithm must be employed. The DPF is one such algorithm. It handles the two nested subproblems using particle filters: subproblem 1 is dealt with using a PF with $N_x$ particles, and subproblem 2 is dealt with using $N_x$ PFs with $N_z$ particles each. The DPF solves these two nested subproblems in 7 steps \cite{chen2011decentralized}, as illustrated in Figure~\ref{fig:dpf}. We only give a brief overview of the main steps of the DPF and state the important equations required for deriving our look-ahead DPF algorithm. We refer the reader to \cite{chen2011decentralized} for all the mathematical details involved in deriving the DPF algorithm and to \cite{doucet2001introduction} for an introduction to particle filtering.
\begin{figure}[t!]
\begin{center}
\par \hspace{-.5mm} \framebox[0.99\linewidth][l]{ 
\begin{minipage}[l]{0.95\linewidth}
\vspace{.3cm}
{\footnotesize
\textbf{Initialize} the particles $\tilde{x}_{0}^{(i)} \sim p(x_{0}), \, i = 1, \dots, N_{x}$, and for each particle  $\tilde{x}_{0}^{(i)}$, the particles $\tilde{z}_{0}^{(i,j)} \sim p(z_{0}|\tilde{x}_{0}^{(i)}),  \, j = 1, \dots, N_{z}$. Initialize $r_1$. \\ \\
\emph{At each time $(t \geq 0)$, perform the following 7 steps:}
\begin{enumerate}
\item \textbf{Measurement update of $x_{0:t}$ given $y_{t}$}. Calculate the importance weights $w_{t}^{(i)} \, , i =1, \dots, N_{z}$ according to: 
\[
w^{(i)}_{t} \propto \dfrac{p_{N_z}(y_{t}|\tilde{x}_{0:t}^{(i)}, y_{0:t-1}) p_{N_z} (\tilde{x}_t^{(i)}|x_{0:t-1}^{(i)}, y_{0:t-1})}{\pi(\tilde{x}_{t}^{(i)}|x_{0:t-1}^{(i)},y_{0:t-1})};  \quad  \sum_{i=1}^{N_{x}} w_{t}^{(i)} = 1 \, .
\]

\item \textbf{Resample} $\{ \tilde{x}_{0:t}^{(i)}, \tilde{z}_{t}^{(i,1)}, \tilde{r}_{t}^{(i,1)}, \dots,  \tilde{z}_{t}^{(i,N_{z})}, \tilde{r}_{t}^{(i,N_{z})} \}, \, i = 1, \dots, N_{x}$  according to $w_{t}^{(i)}$ to generate particles $\{ x_{0:t}^{(i)}, \bar{z}_{t}^{(i,1)}, r_{t}^{(i,1)}, \dots,  \bar{z}_{t}^{(i,N_{z})}, r_{t}^{(i,N_{z})} \}, \, i = 1, \dots, N_{x}$. 

\item \textbf{Measurement update of $z_{t}$ given $y_{t}$}. For $  i = 1, \dots, N_{x}$, the importance weights $\bar{q}_{t}^{(i,j)}$,  $j = 1, \dots, N_{z},$ are evaluated according to: 
\begin{equation}
\bar{q}_{t}^{(i,j)}  \propto p( y_{t} | x_{t}^{(i)}, \bar{z}_{t}^{(i,j)}) r_{t}^{(i,j)};  \quad  \sum_{j=1}^{N_{z}} \bar{q}_{t}^{(i,j)} = 1 \, . \nonumber
\end{equation}

\item \textbf{Propose particles} $\tilde{x}_{t+1}^{(i)}, \, i = 1, \dots, N_{x}$ according to the proposal function $\pi(x_{t+1}|x_{0:t}^{(i)}, y_{0:t})$.

\item \textbf{Measurement update of $z_{t}$ given $\tilde{x}_{t+1}$}. For $ i = 1, \dots, N_{x}$, the importance weights $q_{t}^{(i,j)}, \,\, j= 1, \dots, N_{z}$, are evaluated according to:
\begin{equation}
q_{t}^{(i,j)} \propto p(y_{t} |x_{t}^{(i)}, \bar{z}_{t}^{(i,j)}) p(\tilde{x}_{t+1}^{(i)}|x_{t}^{(i)}, 
\bar{z}_{t}^{(i,j)}) r_{t}^{(i,j)};   \quad \sum_{j=1}^{N_{z}} q_{t}^{(i,j)} = 1 \, .\nonumber
\end{equation}

\item \textbf{Resample} $\bar{z}_{t}^{(i,j)}, i = 1, \dots,N_{x}, j = 1, \dots , N_{z}$ according to  $q_{t}^{(i,j)}$ to obtain $z_t^{(i,j)}$.

\item \textbf{Propose particles} $\tilde{z}_{t+1}^{(i,j)}, i = 1, \dots, N_{x}, j = 1, \dots , N_{z}$  according to the proposal function $\pi(z_{t+1}|\tilde{x}_{0:t+1}^{(i)}, y_{0:t})$. Set $\tilde{x}_{0:t+1}^{(i)} = (x_{0:t}^{(i)}, \tilde{x}_{t+1}^{(i)})$ and compute $\tilde{r}_{t+1}:$
\[
 \tilde{r}_{t+1}^{(i,j)} = \frac{\tilde{p}_{N_{z}} (\tilde{z}_{t+1}^{(i,j)}|\tilde{x}_{0:t+1}^{(i)}, y_{0:t})}{  \pi(\tilde{z}_{t+1}^{(i,j)} | \tilde{x}_{0:t+1}^{(i)}, y_{0:t})} 
\]
\end{enumerate}

\small}\vspace{.2cm}
\end{minipage}}
\end{center}
\vspace{-.1cm}
\caption{{\small \it The DPF algorithm.}}
\label{fig:dpf}
\end{figure}

We describe the 7 steps briefly. Assume that we have Monte Carlo approximations of the distributions of interest from the previous time step:
\begin{eqnarray}
\tilde{p}_{N_x}(x_{0:t-1}|y_{0:t-1}) &=& \frac{1}{N_x} \sum_{i=1}^{N_x} \delta(x_{0:t-1}-x_{0:t-1}^{(i)}) \nonumber \\
{p}_{N_z}(z_{t-1}|x_{0:t-1}^{(i)},y_{0:t-1}) &=&  \sum_{j=1}^{N_z} \bar{q}_{t-1}^{(i,j)}\delta(z_{t-1}-\bar{z}_{t-1}^{(i,j)}), \label{eq:ztm}
\end{eqnarray}
where $\bar{q}_{t-1}^{(i,j)}$ is an importance weight defined in equation~(\ref{eq:qbar}). Assume that we have samples $\tilde{x}_{0:t}^{(i)}|_{i=1}^{N_x}$ from a proposal distribution $\pi(\tilde{x}_{t}|x_{0:t-1}^{(i)},y_{0:t-1})$. Then,  importance sampling enables us to obtain the following approximation of the posterior distribution of $x_{0:t}$:
\[
{p}_{N_x}(x_{0:t}|y_{0:t}) = \sum_{i=1}^{N_x} w_{t}^{(i)}\delta(x_{0:t}-\tilde{x}_{0:t}^{(i)}). 
\]
As in standard particle filtering, the importance weights $w_{t}^{(i)}$ are given by:
\[
w^{(i)}_{t} \propto \dfrac{p_{N_z}(y_{t}|\tilde{x}_{0:t}^{(i)}, y_{0:t-1}) p_{N_z} (\tilde{x}_t^{(i)}|x_{0:t-1}^{(i)}, y_{0:t-1})}{\pi(\tilde{x}_{t}^{(i)}|x_{0:t-1}^{(i)},y_{0:t-1})}.
\]
However, unlike in simple Markov processes, we cannot exploit conditional independence in a trivial manner so as to simplify the numerator. Instead, we express the quantities in the numerator in terms of the following marginals
\begin{eqnarray}
p(x_t|x_{0:t-1}, y_{0:t-1})  &=& \int p(x_t|x_{t-1},z_{t-1}) p(z_{t-1}|x_{0:t-1},y_{0:t-1}) dz_{t-1}
\nonumber\\
p(y_t |{x}_{0:t}, y_{0:t-1}) &=&  \int p(y_t|x_t,z_t) p(z_t|x_{0:t},y_{0:t-1}) dz_{t} 
\nonumber
\end{eqnarray}
and approximate them with the following Monte Carlo estimates:
\begin{eqnarray}
p_{N_z} (\tilde{x}_t|x_{0:t-1}^{(i)}, y_{0:t-1})  &=& \sum_{j=1}^{N_z} \bar{q}_{t-1}^{(i,j)}p(\tilde{x}_{t}|x_{t-1}^{(i)}, \bar{z}_{t-1}^{(i,j)}) \label{eq:approx1} \\
\label{eq:approx2}
p_{N_z} (y_t | \tilde{x}_{0:t}^{(i)}, y_{0:t-1}) &=&    \sum_{j=1}^{N_z} \tilde{r}^{(i,j)}_t p(y_t|\tilde{x}_t, \tilde{z}_t^{(i,j)})/ \sum_{j=1}^{N_z}\tilde{r}^{(i,j)}_t.
\end{eqnarray}
The first expression is a simple Monte Carlo estimate obtained by replacing $p(z_{t-1}|x_{0:t-1},y_{0:t-1})$ with its approximation
${p}_{N_z}(z_{t-1}|x_{0:t-1}^{(i)},y_{0:t-1})$.
In the second expression, we use importance sampling to approximate the integral. In particular, we assume that we have already computed the importance weight
$\tilde{r}^{(i,j)}_t$, defined as follows:
\begin{equation}
 \tilde{r}_{t}^{(i,j)} = \frac{\tilde{p}_{N_{z}} (\tilde{z}_{t}^{(i,j)}|\tilde{x}_{0:t}^{(i)}, y_{0:t-1})}{\pi(\tilde{z}_{t}^{(i,j)} | \tilde{x}_{0:t}^{(i)}, y_{0:t-1})},
\end{equation} 
where $\tilde{z}_{t}^{(i,j)}$ are samples from the proposal mechanism $\pi({z}_{t} | \tilde{x}_{0:t}^{(i)}, y_{0:t-1})$. Note that to compute this importance weight, we require an expression for $\tilde{p}_{N_{z}} (\tilde{z}_{t}^{(i,j)}|\tilde{x}_{0:t}^{(i)}, y_{0:t-1})$. To achieve this, we need to first obtain expressions for $p(z_{t}|x_{0:t},y_{0:t})$ and $p(z_{t}|x_{0:t+1},y_{0:t})$.

Using Bayes rule and conditional independence, we have:
\[
p(z_{t}|x_{0:t}^{(i)},y_{0:t}) \propto p(z_{t}|x_{0:t}^{(i)},y_{0:t-1})p(y_{t}|x_{t}^{(i)},z_{t}).
\]
Since $\pi(z_{t} | \tilde{x}_{0:t}^{(i)}, y_{0:t-1})$ is the proposal mechanism from which the samples $\bar{z}_t^{(i,j)}$ originated, importance sampling yields the following approximation:
\[
{p}_{N_z}(z_{t}|x_{0:t}^{(i)},y_{0:t}) =  \sum_{j=1}^{N_z} \bar{q}_{t}^{(i,j)}\delta(z_{t}-\bar{z}_{t}^{(i,j)}),
\]
where
\begin{equation}
\bar{q}_{t}^{(i,j)} \propto p(y_{t} |x_{t}^{(i)}, \bar{z}_{t}^{(i,j)}) r_{t}^{(i,j)}.
\label{eq:qbar}
\end{equation}
Similarly, by two successive applications of Bayes rule, we have:
\[
p(z_{t}|\tilde{x}_{0:t+1}^{(i)},y_{0:t}) \propto p(z_{t}|x_{0:t}^{(i)},y_{0:t-1})p(y_{t}|x_{t}^{(i)},z_{t})
p(\tilde{x}_{t+1}^{(i)}|x_t^{(i)},z_t).
\]
Using importance sampling, the approximation for this distribution is given by:
\[
{p}_{N_z}(z_{t}|\tilde{x}_{0:t+1}^{(i)},y_{0:t}) =  \sum_{j=1}^{N_z} {q}_{t}^{(i,j)}\delta(z_{t}-\bar{z}_{t}^{(i,j)}),
\]
where
\begin{equation}
{q}_{t}^{(i,j)} \propto p(y_{t} |x_{t}^{(i)}, \bar{z}_{t}^{(i,j)}) p(\tilde{x}_{t+1}^{(i)}|x_t^{(i)},\bar{z}_{t}^{(i,j)}) r_{t}^{(i,j)}.
\label{eq:qnobar}
\end{equation}
Using marginalization and conditional independence, we have
\[
p(z_{t+1}|x_{0:t+1},y_{0:t}) = \int p(z_{t+1}|x_{t:t+1},z_t) p(z_t|x_{0:t+1},y_{0:t}) dz_t.
\]
A Monte Carlo approximation of this quantity results in the expression necessary for computing the numerator of $\tilde{r}_{t+1}:$
\[
\tilde{p}_{N_{z}} ({z}_{t+1}|\tilde{x}_{0:t+1}^{(i)}, y_{0:t})
 = \frac{1}{N_z}\sum_{j=1}^{N_z} p(z_{t+1}|\tilde{x}_{t:t+1}^{(i)}, {z}_{t}^{(i,j)}) 
\]

Note that in contrast to what is done in \cite{chen2011decentralized}, no further approximation of (\ref{eq:approx1}) will be made in the remainder of the derivation. We refer the reader to the results section 
for more details on this.

In the algorithm, shown in Figure 2, the weights $\bar{q}_{t-1}$ and $\tilde{r}_t$ were computed in steps 3 and 7 respectively.
Steps 2 and 6 are standard resampling steps \cite{doucet2001introduction}. 
One has to be careful keeping tracks of indices, tildes and bars, but aside from this, the algorithm follows easily from the standard importance sampling steps for particle filtering.

\subsection{On the choice of proposal distribution} \label{sec:propDist}

A common practice is to use the prior distributions as proposal distributions:  
\[
\pi(x_{t}|x_{0:t-1}^{(i)}, y_{0:t}) =  p(x_{t}|x_{0:t-1}^{(i)}, y_{0:t-1})
\] 
and 
\[
\pi(z_{t}|x_{0:t}^{(i)}, y_{0:t}) = p(z_{t}|x_{0:t}^{(i)}, y_{0:t-1}). 
\]
These proposals are both intuitive and reduce the complexity of the derivations considerably. However, they do not take into account the current observation $y_{t}$. The original DPF algorithm uses these prior proposal distributions. However, a better choice of proposal for $x_t$ that takes into account the current observation $y_t$ is given by:
\[
\pi(x_t|x_{0:t-1}^{(i)}, y_{0:t})  = p(x_t|x_{0:t-1}^{(i)}, y_{0:t}). 
\]
It can be shown that this importance distribution is optimal \cite{doucet2000sequential}. 

Using Bayes rule, the optimal proposal distribution can be written as:
\begin{eqnarray} \label{eq:optdistbayes}
\pi(x_t|x_{0:t-1}^{(i)}, y_{0:t}) &=& \dfrac{p(y_t|x_{0:t-1}^{(i)},x_t,y_{0:t-1})p(x_t|x_{0:t-1}^{(i)},y_{0:t-1})}{p(y_t|y_{0:t-1},x_{0:t-1}^{(i)})}.
\end{eqnarray}
This expression results in the following simplification of the importance weights for $x$:
\begin{eqnarray} \label{eq:weightsopt}
w_t^{(i)} = \dfrac{p(x_{0:t}|y_{0:t})}{\pi(x_{0:t}|y_{0:t})}
		  \propto \dfrac{p(y_{t}|x_{0:t}, y_{0:t-1}) p(x_{t}|x_{0:t-1}, y_{0:t-1})}{\pi(x_{t}|x_{0:t-1},y_{0:t})} = p(y_t |  y_{0:t-1} , x^{(i)}_{0:t-1}). \nonumber
\end{eqnarray}
The predictive distribution $p(y_t |  y_{0:t-1} , x^{(i)}_{0:t-1})$ can be expanded as follows:
\begin{equation}
p(y_t |  y_{0:t-1} , x^{(i)}_{0:t-1})\hspace{0mm}=\hspace{-1mm} \int\hspace{-2mm}\int \hspace{-1mm}  p(y_t | x_t, z_{t}) p(z_{t}| y_{0:t-1} , x^{(i)}_{0:t-1}, x_t) p(x_t |  y_{0:t-1} , x^{(i)}_{0:t-1}) dx_t  dz_{t}  \label{eq:weights2} \nonumber
\end{equation}
Note that $w_{t}^{(i)}$ does not depend on the value of the sample drawn from $\pi(x_t|x_{0:t-1}^{(i)}, y_{0:t})$.

The optimal importance density suffers from two major drawbacks: it requires the ability to sample from  $\pi(x_{t}|x_{0:t-1}^{(i)},y_{0:t})$ and to evaluate the integral over the new states in the calculation of the importance weights.  In general, both of these steps are hard. There are two cases when the use of the optimal importance density is possible: when $\pi(x_{t}|x_{0:t-1}^{(i)},y_{0:t})$ is a member of a finite set, \emph{e.g.} a jump-Markov linear system \cite{doucet2001particle}, or when $\pi(x_{t}|x_{0:t-1}^{(i)},y_{0:t})$ is Gaussian \cite{doucet2000sequential}.

In our case, we can use the Monte Carlo estimates (\ref{eq:approx1}) and (\ref{eq:approx2}) to obtain an approximation of $\pi(x_t|x_{0:t-1}^{(i)}, y_{0:t})$:
\begin{eqnarray}
\hat{\pi}_{N_z}(x_t|x_{0:t-1}^{(i)}, y_{0:t})  &=& \dfrac{{p}_{N_z}(y_t|x_{0:t-1}^{(i)},x_t, y_{0:t-1})   p_{N_z}(x_t|x_{0:t-1}^{(i)}, y_{0:t-1})}{{p}_{N_z}(y_t|y_{0:t-1},x_{0:t-1}^{(i)})} \label{eq:approx3} \\
&=&  \dfrac{\sum_{j=1}^{N_z} r^{(i,j)}_t p(y_t|x_t, z_t^{(i,j)})/ \sum_{j=1}^{N_z} r^{(i,j)}_t  \cdot \sum_{j=1}^{N_z} q_{t-1}^{(i,j)}p(x_{t}|x_{t-1}^{(i)}, z_{t-1}^{(i,j)})}{p_{N_z}(y_t|y_{0:t-1},x_{0:t-1}^{(i)})},  \nonumber 
\end{eqnarray}
where ${p}_{N_z}(y_t|y_{0:t-1},x_{0:t-1}^{(i)})$ is a Monte Carlo approximation of $p(y_t|y_{0:t-1},x_{0:t-1}^{(i)})$, which we derive next.
First, we draw $N_x$ samples $\bar{\bar{x_t}}^{(m)},m=1,\dots,N_x$ using ${p}_{N_z}(x_t|x_{0:t-1}^{(i)}, y_{0:t-1})$. Then, we draw $N_z$ corresponding samples $\bar{\bar{z_t}}^{(m,k)},k=1,\dots,N_z$ according to ${p}_{N_z}(z_{t}| y_{0:t-1} , x^{(i)}_{0:t-1}, \bar{\bar{x}}_t^{(m)})$. Now, the weights can be expanded as follows:
\begin{eqnarray}
 \hat{w}_t^{(i)} & \propto & \dfrac{1}{N_{x}}\dfrac{1}{N_{z}} \sum_{m=1}^{N_x}\sum_{k=1}^{N_z} p(y_t | \bar{\bar{x}}^{(m)}_t, \bar{\bar{z}}_{t}^{(m,k)}). \label{eq:weightsMC}
 \end{eqnarray}
The cost of the algorithm is $N_x \times N_z$. Since $\cal X$ and $\cal Z$ are lower dimensional than ${\cal X}\times{\cal Z}$, the hope is that $N_x \times N_z$ is still lower than the the number of particles $N$ required by a standard particle filter on the joint space ${\cal X}\times{\cal Z}$. Moreover, since typically $N_x$ and $N_z$ are much smaller than $N$, the resampling steps of the DPF are much cheaper than usual $O(N)$ cost of the standard PF. This can therefore result in significant computational gains when parallelizing the algorithm for real-time applications.

\subsection{Look-ahead DPF algorithm}
\begin{figure}[h!]
  \begin{center}
   \includegraphics[width=0.49\textwidth]{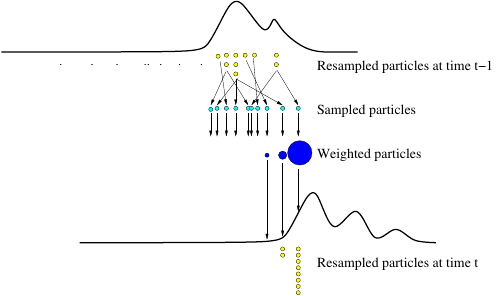}
   \includegraphics[width=0.49\textwidth]{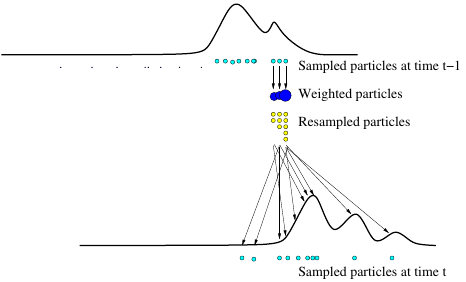}
   \caption{{\small \it PF (left) and look-ahead PF (right) algorithms for a continuous
   one-dimensional problem. For the PF, starting with the
   resampled particles at $t-1$, a new set of particles is proposed at time
   $t$. We compute the importance weight of each particle. Finally, we select the
   fittest particles according to their weights. Note the PF has failed
   to track the two modes appearing on the right of the filtering posterior
   distribution at time $t$. On the other hand,
for the look-ahead filter, we first compute the
   importance weights. After resampling according to these weights, we
   propose new particles at time $t$. With this algorithm, we are more likely
   to propose particles in areas of high probability.}}
   \label{fig:lapf}
  \end{center}
\end{figure}

As mentioned in \cite{df2004fault}, one more improvement is possible when using the optimal proposal distribution. 
The optimal importance weights don't depend on $x_t$ or $z_t$, as we are in fact marginalizing over these variables. Therefore, we can swap the resampling and proposal steps. This enables us to resample (select the fittest) particles at time $t-1$ using information from the future time $t$. We refer the reader to Figure~\ref{fig:lapf} for an intuitive diagram highlighting the benefits of this. Figure~\ref{fig:ladpf} illustrates the 4 steps of the look-ahead DPF algorithm.

\begin{figure}[b!]
\begin{center}
\par \hspace{-.5mm} \framebox[0.99\linewidth][l]{ 
\begin{minipage}[l]{0.95\linewidth}
\vspace{.3cm}
{\footnotesize
\textbf{Initialize} the particles $\tilde{x}_{0}^{(i)} \sim p(x_{0}), \, i = 1, \dots, N_{x}$, and for each particle  $\tilde{x}_{0}^{(i)}$, the particles $\tilde{z}_{0}^{(i,j)} \sim p(z_{0}|\tilde{x}_{0}^{(i)}),  \, j = 1, \dots, N_{z}$. \\ \\
\emph{At each time $(t \geq 0)$, perform the following 4 steps:}
\begin{enumerate}

\item \textbf{Generate $N_x'$ particles} $ \bar{\bar{x}}^{(m)} \sim p_{N_z}(x_t|x_{0:t-1}^{(i)}, y_{0:t-1}), m=1,\dots,N_x'$, and corresponding $N_z'$ particles $\bar{\bar{z_t}}^{(m,k)} \sim p(z_{t}| y_{0:t-1} , x^{(i)}_{0:t-1}, \bar{\bar{x}}_t^{(m)}),k=1,\dots,N_z'$.

\item \textbf{Compute the weights} $w_t^{(i)}$ according to (\ref{eq:weightsMC}).

\item \textbf{Resampling}. Multiply or discard particles $\{\tilde{x}_{0:t-1}^{(i)},\tilde{z}_{0:t-1}^{(i,j)}\}$ using the importance weights $w_t^{(i)}$ to obtain $\{x_{0:t-1}^{(i)},z_{0:t-1}^{(i,j)}\}$.

\item \textbf{Generate new particles}. Obtain $x_t^{(i)} \sim \pi(x_t|x_{0:t-1}^{(i)},y_{0:t})$ and $z_t^{(i,j)} \sim p(z_t|x_{0:t}^{(i)},y_{0:t-1})$.

\end{enumerate}

\small}\vspace{.2cm}
\end{minipage}}
\end{center}
\vspace{-.1cm}
\caption{{\small \it The look-ahead DPF algorithm.}}
\label{fig:ladpf}
\end{figure}

\section{Automatic State Decomposition} \label{sec:EXP3}

The decision of how to decompose the state space of a given system plays a dramatic role in the performance of the DPF. In our setting, with only two nested subproblems, the decomposition problem reduces to deciding in which order the two groups of variables should be sampled. 
This order can have a large impact both on the execution time and on the overall accuracy of the algorithm. If the system is non-stationary, the optimal order can change over time. For this reason, we need to design algorithms to automatically choose the optimal order. 

In this paper, we will adopt classical online bandit algorithms \cite{auer1995gambling} to infer the sampling order. Note that these algorithms are however applicable to the more general problem of choosing state decompositions when splitting the state-space into more than two groups of variables. 

The first algorithm we consider is Hedge (Figure~\ref{fig:hedge}). Hedge chooses action $i$ (out of $K$ possible actions) at time $t$ with probability proportional to exp$(\eta G_i(t-1))$ where $\eta > 0$ is a memory parameter and $G_i(t) = \sum_{t'=1}^{t}r_i(t')$ is the cumulative reward scored by action $i$ from time $1$ to $t$. Actions that repeatedly yield higher rewards quickly gain a higher probability of being selected. In our case, the actions are the different nesting orders. Different reward models are possible. We choose the closeness of the observations to the predicted observations as the reward measure. 
\begin{figure}[t!]
\begin{center}
\par \hspace{-.5mm} \framebox[0.99\linewidth][l]{ 
\begin{minipage}[l]{0.95\linewidth}
\vspace{.2cm}
{\footnotesize
Initialization:
Choose a real number $\eta > 0$. Set $G_i(0) = 0$ for $i=1,\dots,K$.

Repeat for $t=1,2,\dots$:

\begin{enumerate}
\item Choose action $i_t$ according to the distribution:
$p_i(t) = \dfrac{\text{exp}(\eta G_i(t-1))}{\sum_{j=1}^{K}\text{exp}(\eta G_j(t-1))}.$
\item Receive the reward vector $r(t)$ and score the gain $r_{i_t}(t)$.
\item Set $G_i(t) = G_i(t-1) + r_i(t)$ for $i=1,\dots,K$.
\end{enumerate}

\small}\vspace{0cm}
\end{minipage}}
\end{center}
\vspace{-.3cm}
\begin{center}
\par \hspace{-.5mm} \framebox[0.99\linewidth][l]{ 
\begin{minipage}[l]{0.95\linewidth}
\vspace{.2cm}
{\footnotesize
Initialization: Choose $\gamma \in (0,1]$. Initialize Hedge($\eta$).

Repeat for $t=1,2,\dots$:

\begin{enumerate}
\item Get the distribution $p(t)$ from Hedge.
\item Select action $i_t$ to be $j$ with probability $\hat{p}_j(t) = (1 - \gamma)p_j(t) + \dfrac{\gamma}{K}$.
\item Receive reward $r_{i_t}(t) \in [0,1]$.
\item Feed the simulated reward $\hat{r}(t)$ back to Hedge, where 
$
 \hat{r}_j(t) = \left\{ 
  \begin{array}{l l}
   \dfrac{r_{i_t}(t)}{\hat{p}_{i_t}(t)} & \quad \text{if $j = i_t$}\\
    0 & \quad \text{otherwise}\\
  \end{array} \right.
$
\end{enumerate}

\small}\vspace{0cm}
\end{minipage}}
\end{center}
\vspace{-.3cm}
\caption{{\small \it The Hedge (top) and Exp3 (bottom) algorithms \cite{auer1995gambling}.}}
\label{fig:hedge}
\end{figure}
One drawback of Hedge is the fact that each action must be tried to pick the best action (Hedge assumes it has full information about the rewards). This is not ideal since it introduces a large overhead, especially when the number of actions is large. For this reason, we must introduce Exp3 \cite{auer1995gambling}.
Exp3 stands for "Exponential-weight algorithm for Exploration and Exploitation". This algorithm assumes it has only partial information about the reward vector. Exp3 only tries one action at each iteration, and hence it only has information about one reward. Exp3 calls Hedge as a subroutine. For each time interval $t$, Exp3 receives the probability vector $p(t)$ from Hedge, and it selects an action $i_t$ according to a new distribution, $\hat{p}(t)$, which is a mixture of $p(t)$ and the uniform distribution. Using $\hat{p}(t)$ ensures that each action gets tried over time. After receiving the reward $r_{i_t}(t)$ associated with the chosen action, Exp3 must simulate a full reward vector $\hat{r}(t)$ for Hedge. That is, it must fill in the reward for the actions that were not tried (since Hedge requires this information). The pseudo-code is shown in Figure~\ref{fig:hedge}. Both Hedge and Exp3 have vanishing regret \cite{auer1995gambling}.

\section{Simulation Results}

We conduct three experiments. The first experiment compares the performance  
of the look-ahead (LA)-DPF, bootstrap PF and DPF algorithms. To provide a meaningful comparison with results in the literature, we use the model benchmarks adopted in \cite{chen2011decentralized}. These are summarized in Table~\ref{tb:models}. The significant level of nonlinearity, multi-modality and non-stationarity in these models is sufficient to cause classical filtering algorithms, such as the extended Kalman filter, to fail. The second experiment investigates the difference between using a Gaussian approximation to evaluate (\ref{eq:approx1}), as opposed to our Monte Carlo strategy. In the third experiment, we study the behaviour of the automatic state ordering bandit methods.



\begin{table}[h!]
\caption{{\small \it Models used for testing the algorithms.}}
\begin{center}
\begin{tabular}{|@{\hspace{15px}} l @{\hspace{15px}}|@{\hspace{15px}} l |}
\hline
\textbf{Model 1: 2-dimensional example} & \textbf{Model 2: 4-dimensional example} \\
\hline
  &   $x_{1,t+1} = 0.5 x_{1,t} + 8 \sin t+v_t^{x_{1}} $  \\
$x_{t+1} = x_t + \tfrac{z_t}{1+z_t^2}+v_t^x $  &    $x_{2,t+1} = 0.4 x_{1,t} + 0.5 x_{2,t}+v_t^{x_{2}} $ \\
$z_{t+1} = x_t + 0.5z_t + \tfrac{25z_t}{1+z_t^2}$	&  $z_{1,t+1} = z_{1,t}+ \tfrac{z_{2,t}}{1+z_{2,t}^2}+v_t^{z_{1}} $ \\
\hspace{30px}  $+ 8 \cos(1.2t) + v_t^z $ 	& $z_{2,t+1} = z_{1,t+1} + 0.5z_{2,t} + \tfrac{25z_{2,t}}{1+z_{2,t}^2}$    \\
   				&	\hspace{32px} $+ 8 \cos(1.2t) + v_t^{z_{2}}$  \\  
$y_t  =  \arctan(x_t) + \tfrac{z_t^2}{20} + e_t$ 	& $y_{t} = \tfrac{ x_{1,t} + x_{2,t} }{1+x_{1,t}^{2}} + \arctan (z_{1,t}) + \tfrac{z_{2,t}^{2}}{20} + e_{t} $\\
 $[x_0 \, z_0]^T \sim \mathcal{N}(0,I_{2\times2})$ &  $[x_0 \, z_0]^T \sim \mathcal{N}(0,I_{4\times4})$ \\  
$ v_t \sim \mathcal{N}\left(0,\begin{bmatrix}1  & 0.1 \\0.1 & 1 \end{bmatrix}\right)$&   
$ v_t \sim \mathcal{N}\left(0,\begin{bmatrix}1  & 0 & 0 & 0 \\0 & 1 & 0 & 0 \\ 
0& 0 & 1 & 0.1 \\ 0 & 0 & 0.1 & 10 \end{bmatrix}\right) $ \\
 $e_t\sim \mathcal{N}(0,1)$    &  $e_t\sim \mathcal{N}(0,1)$  \\
\hline
 \end{tabular}
 \end{center}
 \label{tb:models}
 \end{table}
 
\subsection{Comparison between LA-DPF, DPF, and PF}

The objective of this experiment is to evaluate the performance of the proposed LA-DPF algorithm with respect to the existing DPF and standard PF algorithms.
The simulations were carried out for 250 time intervals ($t=1,\dots,250$) with a varying number of particles $N_x$ and $N_z$. The accuracy of the state estimate was measured using the root mean square error (RMSE) between the true state and the state estimate. The results were averaged over 500 Monte Carlo simulations. As in \cite{chen2011decentralized}, the number of particles for the regular PF is set to $ N_{x}(N_{z}+1)$ for meaningful comparison.

\renewcommand{\thesubfigure}{\arabic{subfigure}}
\begin{figure}[t!]
\centering
\includegraphics[width=0.9\textwidth]{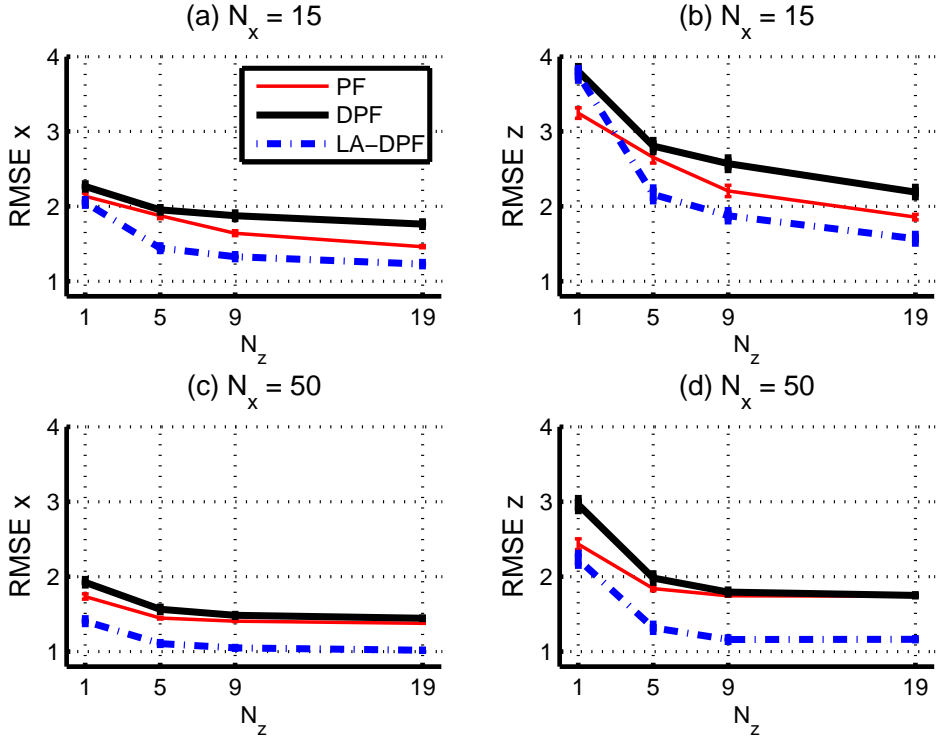}
 \caption{{\small \it Comparison among the PF, DPF and LA-DPF algorithms for model 1. The plots show the RMSE (mean and error bars) for the states $x$ and $z$ respectively as a function of the number of particles. Note that even in a sequential implementation, LA-DPF outperforms the standard PF in terms of this measure.}}
\label{fig:plotsRMSE2d}
\end{figure}
\begin{figure}[h!]
\centering
\subfigure{
\includegraphics[width=0.6\textwidth]{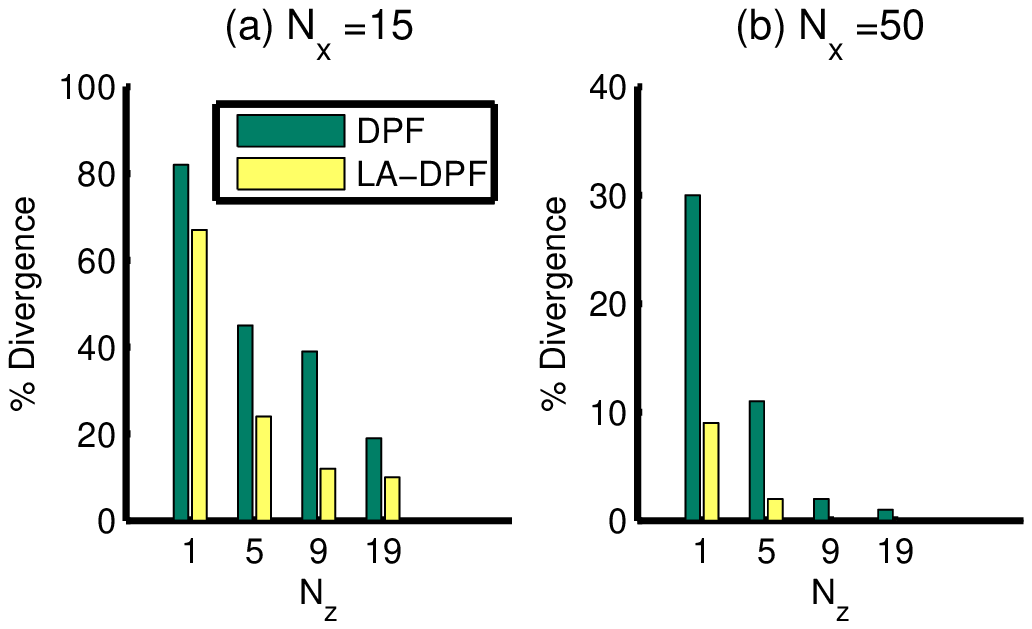} }
\subfigure{
\includegraphics[width=0.35\textwidth]{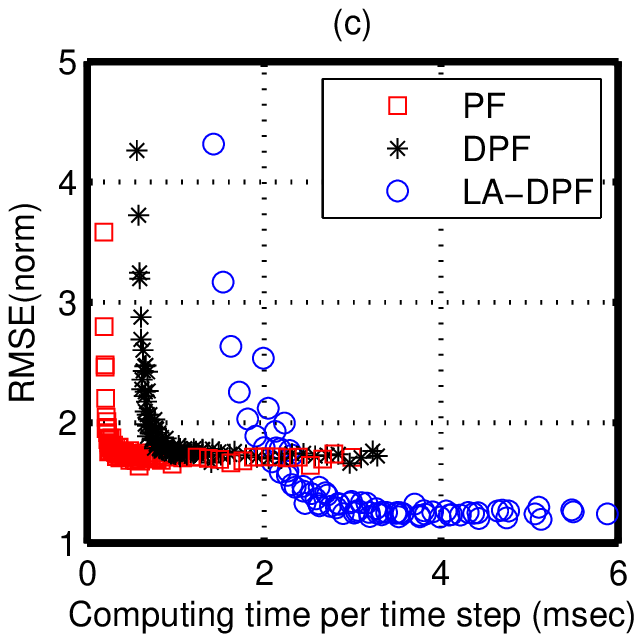} }

 \caption{{\small \it Comparison of LA-DPF against DPF in terms of $\{$(a),(b)$\}$ divergence rate for different numbers of particles and (c) attained RMSE as a function of computing power.}}
\label{fig:plotsCP2d}
\end{figure}

\begin{figure}[t!]
\centering
\includegraphics[width=\textwidth]{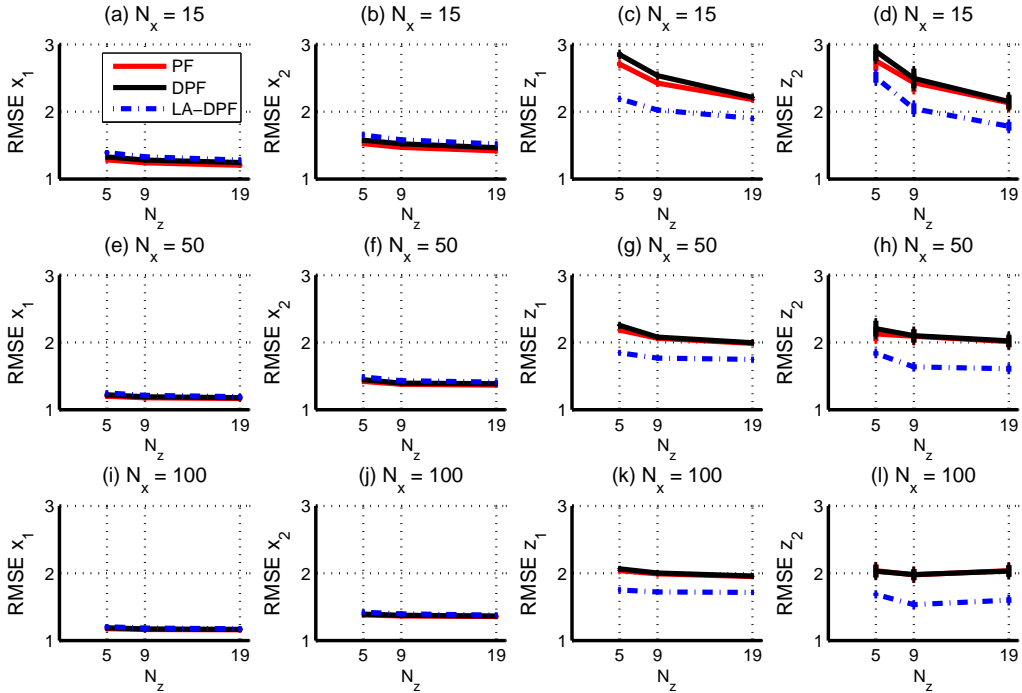}
 \caption{{\small \it Comparison among the PF, DPF and LA-DPF algorithms for model 2. The plots show the RMSE (mean and error bars) for the states $x_1$, $x_2$, $z_1$ and $z_2$ as a function of the number of particles. Note that LA-DPF outperforms both DPF and PF.}}
\label{fig:plotsRMSE4d}
\end{figure}

\begin{figure}[t!]
\centering
\subfigure{
\includegraphics[width=0.66\textwidth]{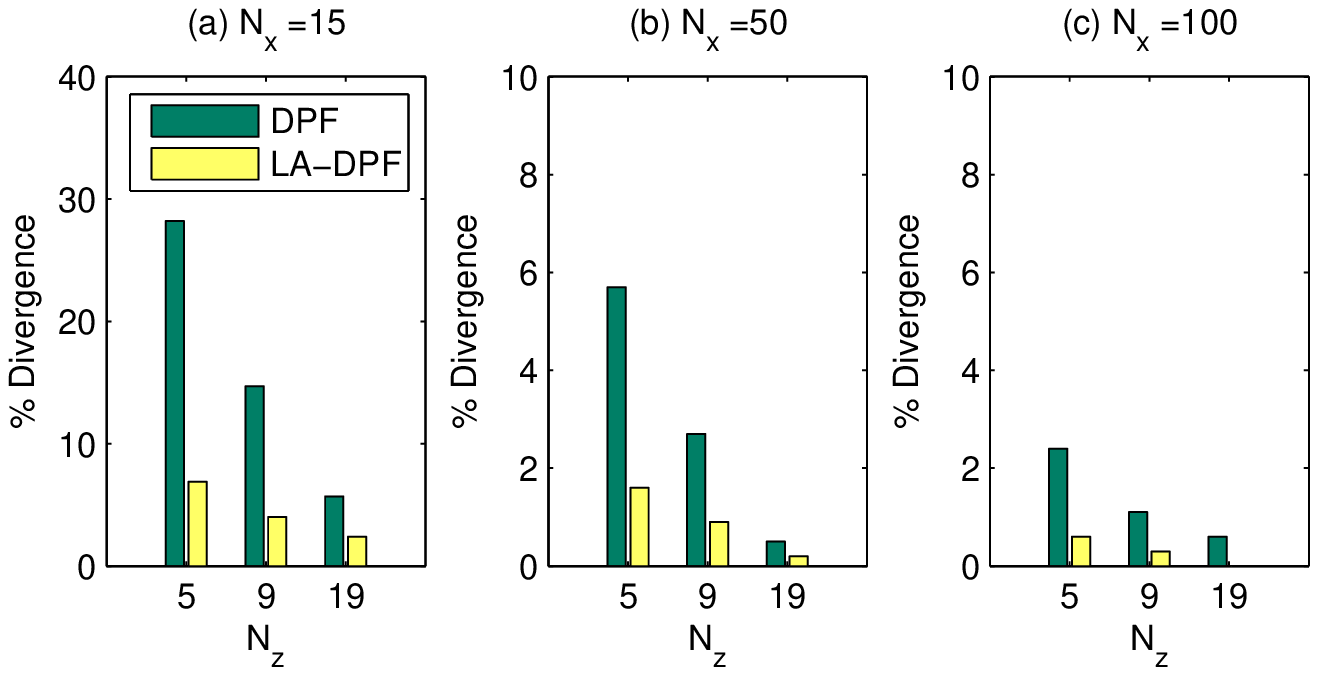} }
\subfigure{
\includegraphics[width=0.3\textwidth]{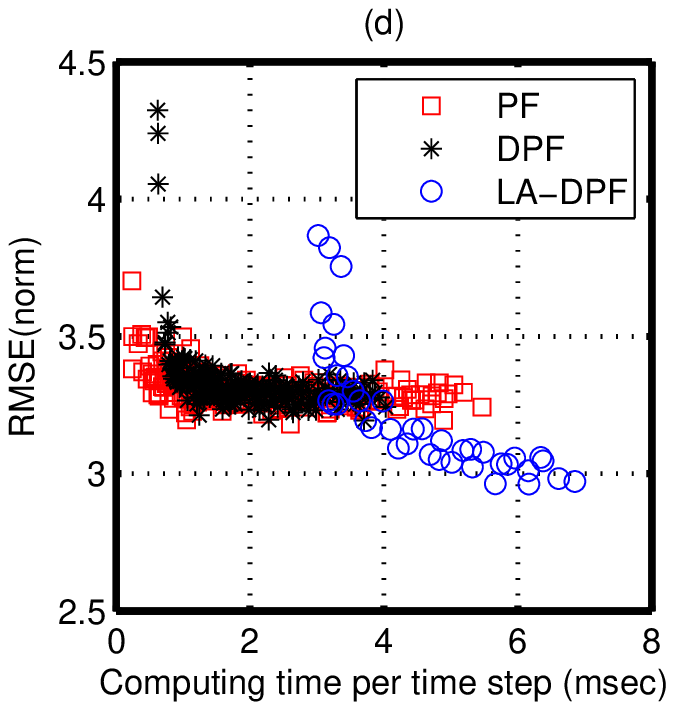} }
\caption{{\small \it The first 3 columns show the divergence plots for model 2. Again, for any number of particles the DPF has a higher likelihood of diverging than the LA-DPF. The rightmost column compares the LA-DPF against the PF and DPF in terms of attained RMSE as a function of computing power.}}
\label{fig:plotsCP4d}
\end{figure}

The results are summarized in Figures~\ref{fig:plotsRMSE2d} to \ref{fig:plotsCP4d}. Figure~\ref{fig:plotsRMSE2d} shows the RMSE (mean and 90\% error bars) for the $x$ and $z$ states for different numbers of particles with model 1. It is clear from these plots that although PF can have lower error than DPF in a serial implementation of the algorithms (as also reported in \cite{chen2011decentralized}), the same is not true for LA-DPF. LA-DPF does significantly better as a function of the number of particles. Plots (a) and (b) of Figure~\ref{fig:plotsCP2d} illustrate that for the same number of particles, the divergence rate of the DPF is higher than the one for LA-DPF. The difference is more significant for higher $N_x$. All methods work well for a reasonable number of particles. 
Figure~\ref{fig:plotsCP2d} (c) is a scatter plot for runs with multiple random $N_x$ and $N_z$ values. It illustrates the accuracy versus computation time trade-off for DPF and LA-DPF. The latter can attain much lower error than the PF and DPF variants. We should point out that both DPF and LA-DPF can equally benefit from parallel implementation as they follow the same state-decomposition strategy to reduce the cost of resampling. We should also note that for some transition models, the execution time of LA-DPF could be decreased even further using kd-trees as proposed in \cite{klaas4toward}.


Figures~\ref{fig:plotsRMSE4d} and \ref{fig:plotsCP4d} are the results for Model 2, and are analogous to Figures~~\ref{fig:plotsRMSE2d} and \ref{fig:plotsCP2d}, respectively. Again, it can be seen that LA-DPF performs significantly better than PF and DPF (or at worst, the performance is equivalent). Also, the divergence rate is much smaller for the LA-DPF than for the DPF. Similar conclusions to the ones for Model 1 can be drawn when it comes to the accuracy versus computation trade-off.

\subsection{Comparison between DPF with Gaussian and Monte Carlo approximations}

We note that in \cite{chen2011decentralized}, a Gaussian approximation is used to estimate the proposal distribution (\ref{eq:approx1}). It is however possible to use Monte Carlo approximations instead of Gaussian approximations.
We carried experiments to verify that both approaches yield similar results for the model of Table \ref{tb:models}. We ran the DPF with Gaussian approximation and the DPF with Monte Carlo approximation to compare their respective RMSEs. The experiment was performed 100 times, with $N_{x} = 100$ and $N_{z} = 19$. Table \ref{tb:GaussianMCcomp} shows the results.
\begin{table}[t!]
\caption{{\small \it Average RMSE for DPF (Gaussian) and DPF (Monte Carlo) algorithms, $N_{x} = 100$, $N_{z} = 19$ (100 runs)}}
\begin{center}
\begin{tabular}{c c c}
\toprule
					     &  Gaussian approximation			& 		 Monte Carlo approximation	\\
 \hline 
  $\text{RMSE}_{x}$            &        1.3197     &  1.3216 \\ 
  $\text{RMSE}_{z}$        &   1.1705   & 1.1640 \\
\bottomrule
\end{tabular}
\end{center}
\label{tb:GaussianMCcomp}
\end{table}
The results in terms of the RMSE are equivalent (within 1\% of one another). However, the Monte Carlo method has the advantage of being universal in the sense that it does not require any assumption on the types of distributions that we are dealing with. It should be expected that for non-standard noise models, Gaussian approximation could result in very poor performance.

%

\subsection{Automatic state decomposition for the two dimensional example}

In this experiment, we use Hedge and Exp3 to choose the optimal sampling order among two actions (action 1= sampling $x$ first, then $z$, action 2 = sampling $z$ first, then $x$). We tried each action separately and verified that this ordering has a large effect on the RMSE. 

We designed the reward function to be
$r_{i_{t}}= \tfrac{\alpha}{\alpha+\epsilon_{t}^{2}} (0<r_{i_{t}}\leq1)$, where $\alpha$ is a small number (say 0.001) and $\epsilon_{t}$ is the difference between the one-step-ahead predicted observation and the actual observation. Other functionals of $\epsilon$ could also work just as easily.

\begin{figure}[h!]
\centering
  \begin{minipage}[c]{0.49\textwidth}
   	 \includegraphics[width=0.99\textwidth, height = 110px]{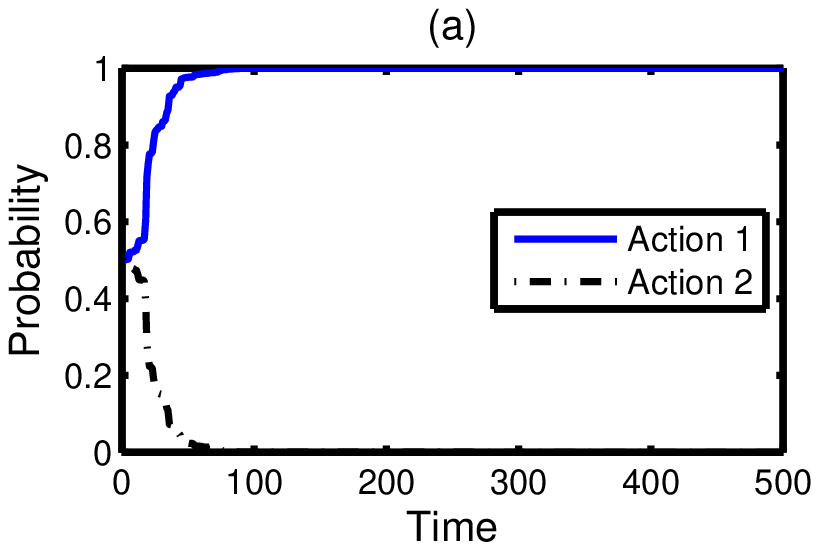}
   \end{minipage}
  \begin{minipage}[c]{0.49\textwidth}
   	 \includegraphics[width=0.99\textwidth, height = 110px]{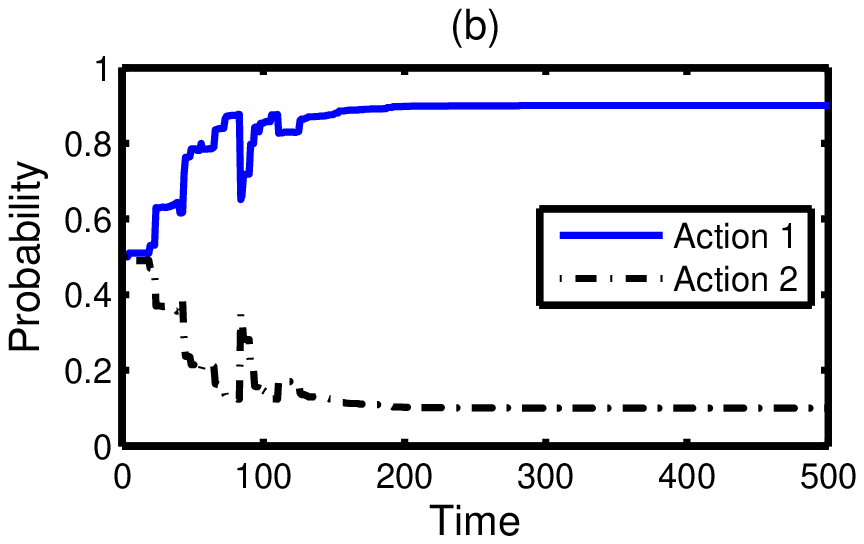}
   \end{minipage}
\caption {{\small \it Automatic state decomposition using (a) Hedge and (b) Exp3.}}
\label{fig:HedgeExp3}
\end{figure}

Figure~\ref{fig:HedgeExp3} shows the evolution of the probability for the two actions as a function of time for $\gamma=0.2$ and $\eta= 0.5$ for both algorithms. The algorithms are able to converge to the action with the lowest RMSE. As expected, Hedge converges faster and more smoothly than Exp3. However, as the state space grows, Hedge becomes much less efficient than Exp3 because it relies on trying every action at each time step. Therefore, Exp3 is preferred.

In another experiment, we introduce a change point at $t=600$, so as to also assess whether the bandit algorithm can adapt. For the change-point, we simply swap the transition models for model 1. We observed the same behavior with model 2.

Figure~\ref{fig:exp3} (a) shows the evolution of the probability for the two actions as a function of time for $\gamma=0.2$ and $\eta= 0.5$. The algorithm is able to converge to the action with the lowest RMSE. After the change-point, the algorithm is able to gradually adapt. The rate of this adaptation is controlled by the hyper-parameters of the control algorithm. It should be noted that we chose a very dramatic artificial change-point, which perturbs stability of the particle filters significantly. In practical applications we would expect more gradual model drift and, hence, better adaptation.
Plot (b) shows the action sequence for the last 100 time steps and Table (c) compares the RMSE values incurred by Exp3 and the fixed action policies for the last 1000 time steps. As expected, Exp3 achieves a better result than the worst action. If we know our setting is stationary, then we can stop adapting and attain the same RMSE as the best action.

\addtocounter{figure}{-1}

\begin{figure}[htb]
  \centering
  \begin{minipage}[c]{0.78\textwidth}
   	 \includegraphics[width=0.99\textwidth, height = 110px]{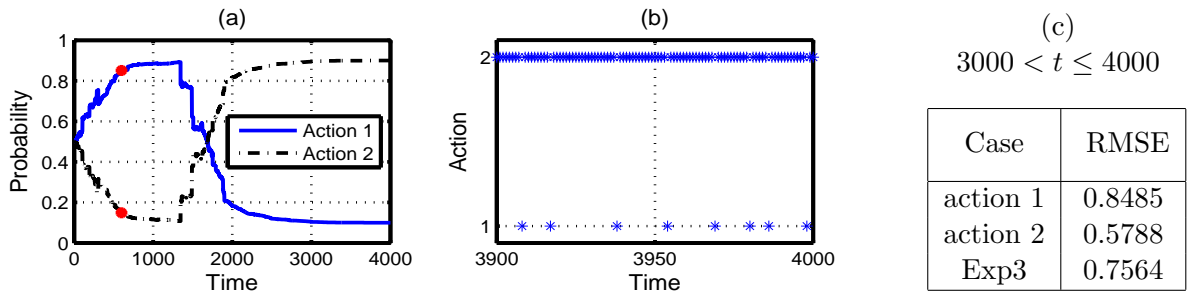}
   \end{minipage}
   \begin{minipage}[r]{0.21\textwidth}
   \captionsetup{labelformat=empty}
   \caption{(c) \\ $ 3000<t \leq 4000$}
   \begin{tabular}{|c |c|}
   \hline 
  \multirow{2}{*}{Case} 			& \multirow{2}{*}{ RMSE}\\
   				&   \\			
   \hline
   				
   action 1			 & 	0.8485	\\
    action 2			& 	0.5788	\\
    Exp3			&	0.7564	\\
   \hline  
    \end{tabular}
  \end{minipage}
 \caption {{\small \it Automatic state decomposition using Exp3 when introducing a change point. a) Evolution of probability of each action, where the red dot indicates the time step at which we switch the models, b) Action sequence for the last 100 time steps, c) Comparison of the RMSE for action 1, action 2, and Exp3 for the last 1000 time steps.}}
 \label{fig:exp3}
  \end{figure}

\section{Concluding Remarks and Future Work}

The DPF algorithm is a new ingenious particle filter that holds great promise. In this paper, we proposed two algorithmic improvements: a look-ahead formulation and an automatic state-decomposition strategy. The look-ahead strategy performed remarkably well. Even though the original motivation for the DPF was to improve the parallelization level of particle filters, our experiments show that the look-ahead strategy works better than the widely used PF algorithm even in a serial implementation. 

In the experiments we also assessed the performance of the state-decomposition strategy using Exp3. The simple demonstration made it clear that it is possible to use bandits to automatically configure the filter. However, we also must point out that this set up was simple enough that Exp3 could handle it. As we move on to more sophisticated partitioning schemes, it will become necessary to adopt more powerful control strategies using correlated bandit strategies or Bayesian optimization; see for example \cite{Li2010ac,brochu2010tutorial,Srinivas2010,Lizotte2011,Hutter:2009a}. 

The immediate future work directions are to test the look-ahead DPF on practical settings and to carry out an empirical evaluation using GPUs. A longer term goal is to increase the level of partitioning (having more than two levels of nesting) of the state space. How such a strategy behaves in high-dimensions is of great interest. Another long term goal is to capitalize on the ideas proposed here to distribute the observations across multiple cores. That is, both the states and the observations should be decomposed for greater applicability to vast streaming datasets.

\section*{Acknowledgements}

We would like to thank Arnaud Doucet, Alex Smola and Anthony Lee for useful discussions on this topic, which to a large extent shaped this paper. This work was supported by NSERC.

\bibliographystyle{unsrt}    
\bibliography{pfref}       

\end{document}